%% file: root.tex
\documentclass[letterpaper, 10 pt, conference]{ieeeconf}  

\IEEEoverridecommandlockouts                              

\usepackage{times} 
\usepackage{amsmath} 
\usepackage{amssymb}  
\usepackage{epsfig} 
\usepackage{amsfonts}
\usepackage{bm}
\usepackage{adjustbox}
\usepackage{multirow}
\usepackage{wrapfig}
\usepackage{caption}
\usepackage{subcaption}
\usepackage{mathtools}
\usepackage{url}
\usepackage{hyperref}
\usepackage{booktabs}       
\usepackage{pifont}
\usepackage{nicefrac}
\usepackage{lipsum}
\usepackage{cuted}

\usepackage{graphicx}

\DeclareMathOperator*{\argmin}{\arg\!\min}

\title{\input{includes/title}}

\author{\input{includes/authors.tex}}

\begin{document}

\maketitle
\thispagestyle{empty}
\pagestyle{empty}


\begin{abstract}
\input{includes/abstract}
\end{abstract}



\input{includes/sec1-intro}

\input{includes/sec2-relatedwork}

\input{includes/sec3-method}

\input{includes/sec4-exp}

\input{includes/sec5-conclusion}



\noindent \textbf{Acknowledgment.} This work was supported by the Sony Research Award Program and the National Science Foundation (NSF) under Grant No. 2346528.

\bibliographystyle{IEEEtran}
\bibliography{references}

\end{document}

%% file: includes/title.tex
{\LARGE \bf RobotFingerPrint: Unified Gripper Coordinate Space for \\ Multi-Gripper Grasp Synthesis and Transfer}

%% file: includes/authors.tex
Ninad Khargonkar, Luis Felipe Casas, Balakrishnan Prabhakaran, Yu Xiang
\thanks{All authors are affiliated with the Department of Computer Science, University of Texas at Dallas, Richardson, TX 75080, USA {\tt\footnotesize \{ninadarun.khargonkar, luis.casasmurillo, bprabhakaran, yu.xiang\}@utdallas.edu}}%


%% file: includes/abstract.tex
We introduce a novel grasp representation named the Unified Gripper Coordinate Space (UGCS) for grasp synthesis and grasp transfer. Our representation leverages spherical coordinates to create a shared coordinate space across different robot grippers, enabling it to synthesize and transfer grasps for both novel objects and previously unseen grippers. The strength of this representation lies in the ability to map palm and fingers of a gripper and the unified coordinate space. Grasp synthesis is formulated as predicting the unified spherical coordinates on object surface points via a conditional variational autoencoder. The predicted unified gripper coordinates establish exact correspondences between the gripper and object points, which is used to optimize grasp pose and joint values. Grasp transfer is facilitated through the point-to-point correspondence between any two (potentially unseen) grippers and solved via a similar optimization. Extensive simulation and real-world experiments showcase the efficacy of the unified grasp representation for grasp synthesis in generating stable and diverse grasps. Similarly, we showcase real-world grasp transfer from human demonstrations across different objects.\footnote{Project page with code and videos is available at: \url{https://irvlutd.github.io/RobotFingerPrint}}

%% file: includes/sec1-intro.tex
\section{INTRODUCTION}
\label{sec:intro}
Grasp synthesis is a fundamental problem in robot manipulation. Traditional grasp planners such as~GraspIt!~\cite{miller2004graspit} and OpenRAVE~\cite{diankov2008openrave} adopt a sampling and evaluation strategy, in which the grasps are sampled around an object and a force closure analysis~\cite{nguyen1988constructing} is used to evaluate the grasps. These grasp planners are usually computationally expensive and may require users to manually specify contact points on the grippers. Recently, learning-based grasp synthesis has received more attention for generalization to objects and grippers. These methods use generative machine learning models such as variational autoencoders~\cite{mousavian20196,li2023gendexgrasp} or diffusion models~\cite{urain2023se} for grasp generation. They are trained on large-scale grasp datasets with various objects, which enable them to predict grasps for unseen objects~\cite{eppner2021acronym,li2023gendexgrasp,turpin2023fast-graspd}. However, most learning-based models are trained only for a specific gripper~\cite{mousavian20196,sundermeyer2021contact, mayer2022ffhnet}. Consequently, these models lack the ability to synthesize grasps for robotic hands outside their training domain.

In order to train a single grasp synthesis model to support multiple grippers, researchers have explored predicting intermediate grasp representations instead of directly predicting the grasp pose and joint values which are specific for each gripper. In the literature, two types of intermediate grasp representation have been developed: \emph{gripper keypoints} and \emph{contact maps}. For example, UniGrasp~\cite{shao2020unigrasp} and GeoMatch~\cite{geomatch} train neural networks to predict the locations of gripper keypoints on an input object and then solve inverse kinematics to obtain the grasps. Furthermore, GenDexGrasp~\cite{li2023gendexgrasp} predicts a contact map of a given object and then solves an optimization problem to generate grasps that comply with the generated contact map. These intermediate grasps representations have been used to synthesize grasps for various grippers, even for unseen grippers outside the training domain.

\begin{figure}
\centering
\begin{center} 
\includegraphics[width=0.8\linewidth]{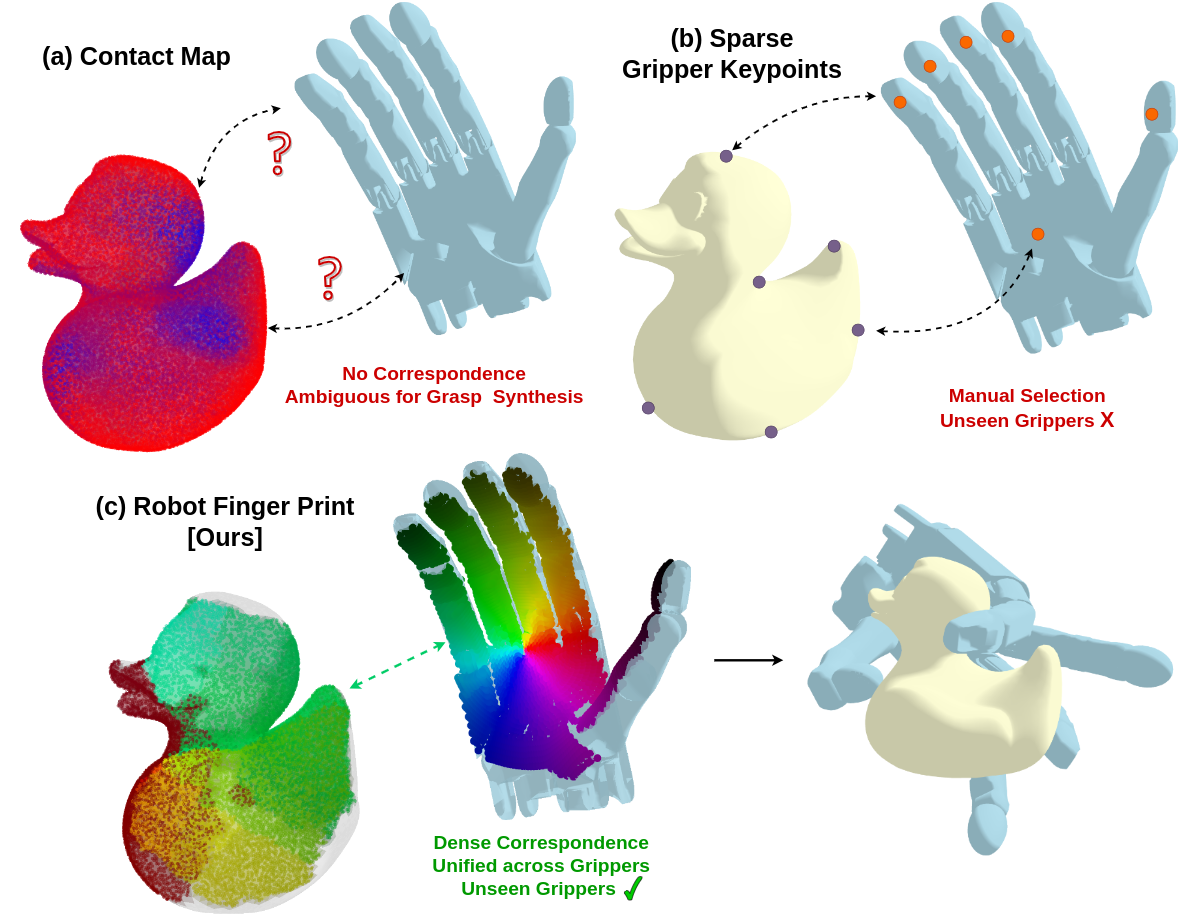}
\caption{Comparison between three different representations for grasp synthesis: (a) contact map; (b) sparse gripper keypoints; (c) our unified gripper coordinate space}
\label{fig:teaser}
\vspace{-7mm}
\end{center}
\end{figure}

We illustrate the limitations in both gripper keypoints and contact maps for grasp synthesis in Fig.~\ref{fig:teaser}. On one hand, gripper keypoints are sparse. Therefore, noise for any predicted keypoint location can significantly affect grasp quality. Users must manually specify these keypoints~\cite{geomatch}, and the keypoint prediction order is predetermined, thus inhibiting flexibility. Similarly, contact maps are ambiguous for grasp synthesis when mapping grippers to contact locations. They only provide information about the distances between the gripper and the object, but there is no information on how and where the gripper fingers should touch the object~\cite{li2023gendexgrasp}. Consequently, these representations are neither suitable for transferring one grasp from one gripper to another gripper.


Motivated by the normalized object coordinate space in object pose estimation~\cite{wang2019normalized} and dense human pose estimation~\cite{guler2018densepose}, we introduce a novel intermediate grasp representation named the \emph{Unified Gripper Coordinate Space (UGCS)}. This space, denoted as $S^2$, is the two-dimensional surface of a sphere in a three-dimensional space. Any point within the UGCS is defined by its longitude $\lambda \in [-\pi, \pi)$ and its latitude $\varphi \in [-\frac{\pi}{2},  \frac{\pi}{2})$. This space is intended to be shared between all robotic hands and used to map the inner surface of the grippers (gripper ``palm''). Here, every 3D point on the palm is assigned a 2D coordinate $(\lambda, \varphi)$. We refer to these points as the ``finger print'' of the robotic gripper. For grasp synthesis, we train a Conditional Variational Auto-Encoder (CVAE) to map a latent vector and a point cloud of an object to the UGCS. Specifically, CVAE predicts the UGCS points $(\lambda, \varphi)$ for every point on the surface of the object. Then, an objective function is optimized using the predicted coordinates to obtain the synthesized grasp configuration. Cross-gripper grasp transfer is computed by first initializing a mutual correspondence between surface points for source and target grippers, via their respective coordinate maps. Then the target gripper grasp is inferred from an optimization that tries to minimize the distance between the corresponding points.

Compared to keypoints and contact maps as grasp representations, our proposed UGCS establishes dense correspondences between a gripper and an object. 
We conducted grasp synthesis experiments using the MultiDex dataset~\cite{li2023gendexgrasp} and compared it to methods that use alternative intermediate grasp representations~\cite{li2023gendexgrasp, geomatch}. Our method, RobotFingerPrint (RFP), was found to compare favorably against prior works
in terms of both grasp success rate and grasp diversity. Furthermore, we tested our method with out-of-domain Fetch robot~\cite{wise2016fetch} gripper by performing real-world manipulation on the SceneReplica~\cite{khargonkar2024scenereplica} benchmark. The grasps generated by the RFP were able to outperform on the model-based grasp scenario with the GraspIt~\cite{miller2004graspit} baseline of the benchmark. 
We also showed the effectiveness in grasp transfer from human hand to the Fetch gripper. Such a transfer pipeline can be an effective tool in learning from demonstration scenarios with purely visual information.
Our contributions are summarized as follows.
\begin{itemize}
    \item We introduce the unified gripper coordinate space that establishes dense correspondences across different grippers via a novel shared representation space. 
    \item Diverse and stable object-centric grasp synthesis across in/out-domain grippers and object sets with stronger pose initialization.   
    \item Generalized grasp transfer across different grippers without requiring manual re-targeting. 
\end{itemize}

%% file: includes/sec2-relatedwork.tex
\section{Related Work}
\label{sec:related-work}

\subsection{Analytic Grasp Synthesis}   
Classical grasp planners such as GraspIt!~\cite{miller2004graspit} and OpenRAVE grasp planner~\cite{diankov2008openrave} use analytic approaches to access grasp quality. Such methods employ task wrench space analysis~\cite{ferrari1992planning,borst2004grasp} or force closure analysis~\cite{nguyen1988constructing,berenson2008grasp} to assess the feasibility of a proposed grasp. These traditional grasp synthesis methods usually use sampling-based optimization algorithms, such as simulated annealing to sample grasps, and then evaluate them, which makes the optimization quite slow. Some methods also require the user to manually specify the preferred contact regions on the gripper surface prior to starting the grasp generation process. Recent analytic grasp synthesis leverages Differentiable Force Closure (DFC)~\cite{liu2021synthesizing} in grasp optimization, where the optimization runtime is significantly improved. The main limitation of analytic methods is that the success rate of grasping may not be guaranteed because force closure analysis cannot account for other factors in grasping such as friction, materials of objects and noises in control.

\subsection{Learning-based Grasp Synthesis}
To improve speed and success in grasp synthesis, recent efforts have been devoted to learning-based grasp synthesis. Datasets have been designed for machine learning methods on grasping via grasp stability verification in physics simulators. Representative datasets include Acronym~\cite{eppner2021acronym}, UniGrasp~\cite{shao2020unigrasp}, DexGraspNet~\cite{wang2023dexgraspnet}, Fast-Grasp'D~\cite{turpin2023fast-graspd}, GenDexGrasp~\cite{li2023gendexgrasp} and MultiGripperGrasp~\cite{casas2024multigrippergrasp}. These datasets contain different numbers of grippers, objects, and grasps, which can be used to learn generative models for grasp synthesis such as variational auto-encoders or diffusion models.

\subsubsection{Single-Gripper Models}
Most learning-based methods can only plan grasps for a specific type of robotic gripper. For example, \cite{mousavian20196,fang2020graspnet1billion,sundermeyer2021contact} deal with parallel jaw grippers while FFHNet~\cite{mayer2022ffhnet} is designed for a five-finger gripper. Unlike classical approaches, these learning-based methods cannot plan grasps for a variety of robotic hands.

\subsubsection{Multi-Gripper Models}
In recent years, a few approaches have been proposed for grasp synthesis of multiple grippers. For example, AdaGrasp~\cite{xu2021adagrasp} convolves over both normalized gripper features and scene features to predict grasps, which in turn enables it to generalize over robotic hands. UniGrasp~\cite{shao2020unigrasp} trains a neural network to predict finger-tip contact points on objects, and then solves inverse kinematics to generate grasps. However, both AdaGrasp and UniGrasp cannot scale-up to grippers with more than 3 fingers. On the other hand, GenDexGrasp~\cite{li2023gendexgrasp} use a CVAE to predict contact maps on a given object and then solve an optimization problem using the predicted contact map to generate a grasp. GeoMatch~\cite{geomatch} follows a similar idea as UniGrasp~\cite{shao2020unigrasp} to predict keypoints of grippers on object surfaces and then solve inverse kinematics to compute grasps. We introduce a new method for multi-gripper grasp synthesis based on our UGCS representation.

\subsection{Grasp Transfer}

For grasp transfer, previous work either focuses on the problem of cross-object transfer~\cite{wu2024cross, madry2012object} or formulates it as retrieval from an offline dataset~\cite{khargonkar2022neuralgrasps} that inhibits generalization to unseen objects and grippers.
Unlike existing methods, we introduce a novel grasp representation that encodes dense correspondences of gripper surfaces and object surfaces. This representation enables grasp transfer that generalizes to novel grippers and objects.

%% file: includes/sec3-method.tex

\begin{figure*}[!h]
    \centering
    \begin{center} 
    \includegraphics[width=0.8\textwidth]{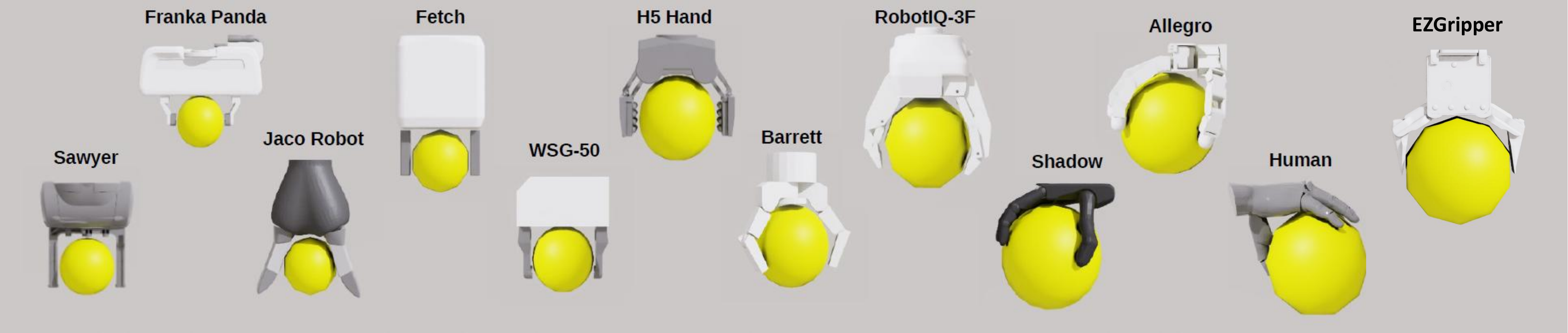}
    \caption{Max spheres across all the 12 different grippers including the gripper set from the MultiGripperGrasp~\cite{casas2024multigrippergrasp} dataset.}
    \label{fig:max_spheres}
    \vspace{-6mm}
    \end{center}
\end{figure*}
\label{sec:method}

\section{Unified Gripper Coordinate Space}\label{sec:method-ugcs}
The Unified Gripper Coordinate Space (UGCS) is a coordinate space that maps the points on the interior surface of a gripper onto the surface of a sphere. This coordinate space is common to different grippers and provides a dense representation of the interior surface. 
A sphere was chosen because of its smooth surface point representation and the fact that similar contact regions can be marked unambiguously across grippers with varying scales. The UGCS representation assigns a normalized coordinate $(\lambda, \varphi)$, with each component in $[0, 1]$ obtained after normalizing the longitude and latitude in their ranges $\lambda \in [-\pi, \pi)$ and $\varphi \in [-\frac{\pi}{2},  \frac{\pi}{2})$. Next, we present the method for constructing such a unified coordinate space across any gripper via their grasp on an associated maximal sphere.

\begin{figure}[!h]
    \centering
    \begin{center} 
    \includegraphics[width=0.8\linewidth]{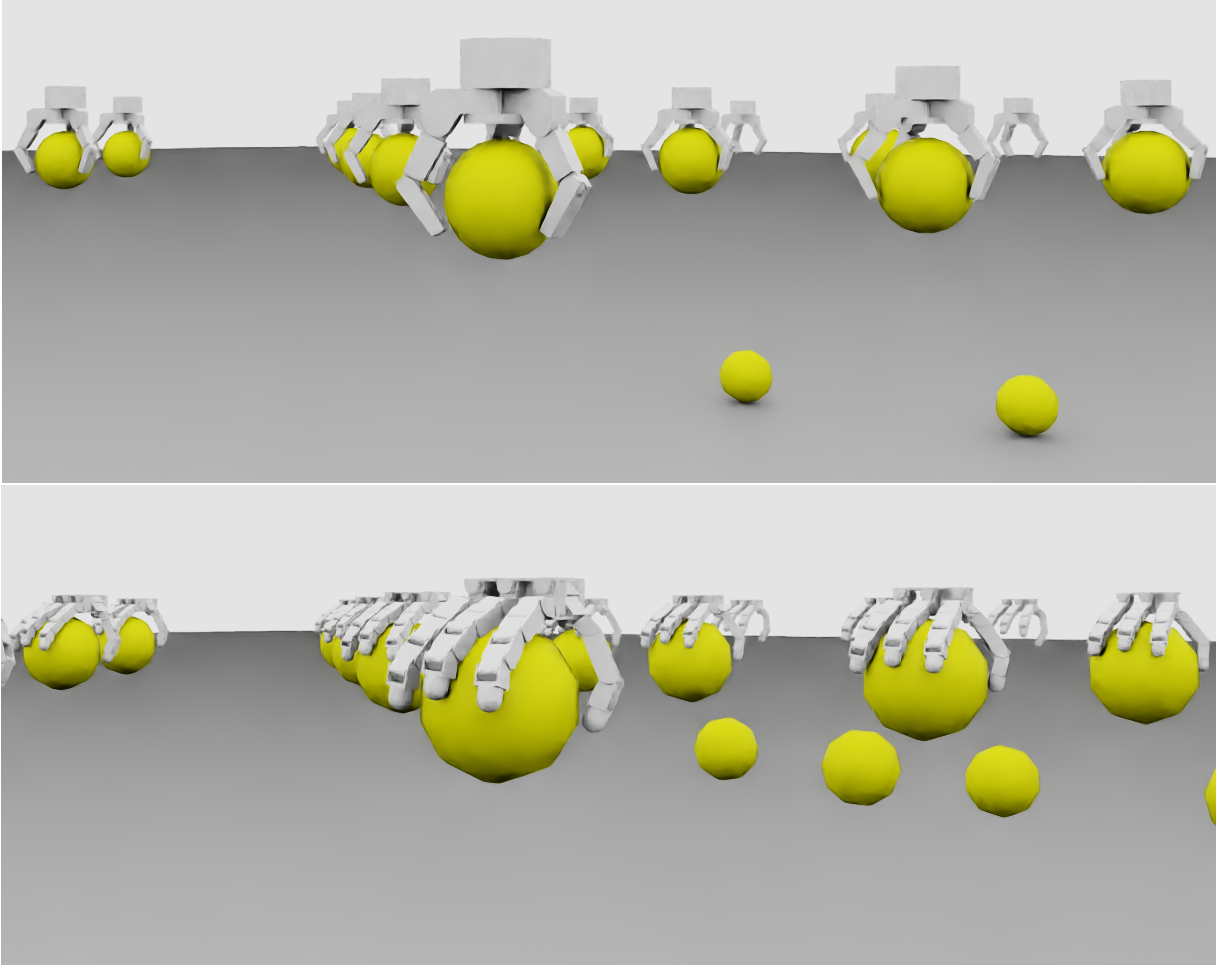}
    \caption{Maximal Graspable Sphere: We test for multiple spheres with varying radii in parallel using Isaac Sim~\cite{nvidia2023-isaac-sim}}
    \label{fig:max_sphere_exp}
    \vspace{-6mm}
    \end{center}
\end{figure}

\subsection{Maximal Sphere Computation}
Since the geometry and kinematics between grippers differ significantly, there is no single sphere that can be used to assign this coordinate mapping. Hence, we created a test bed in simulation to obtain the maximal graspable sphere of each gripper for assigning the coordinate mapping. We used the NVIDIA Isaac Sim \cite{nvidia2023-isaac-sim} robotic simulator where the simulation consisted of top-down grasps of spheres with increasing radii. These spheres were initialized with a common pole position and each gripper performed an encompassing grasp from the south pole.
Each grasp was simulated for 3 seconds after the gripper came in contact with the sphere. 
The maximal spheres of grippers in the MultiGripperGrasp \cite{casas2024multigrippergrasp} and MultiDex \cite{li2023gendexgrasp} datasets can be seen in Fig.~\ref{fig:max_spheres}, while Fig.~\ref{fig:max_sphere_exp} shows the process of sphere-grasping tests in Isaac Sim. 

\subsection{UGCS for Gripper Points}\label{sec:method-ugcs-gripper-pts}
\begin{figure*}[!ht]
    \centering
    \begin{center}
    \includegraphics[width=0.8\textwidth]{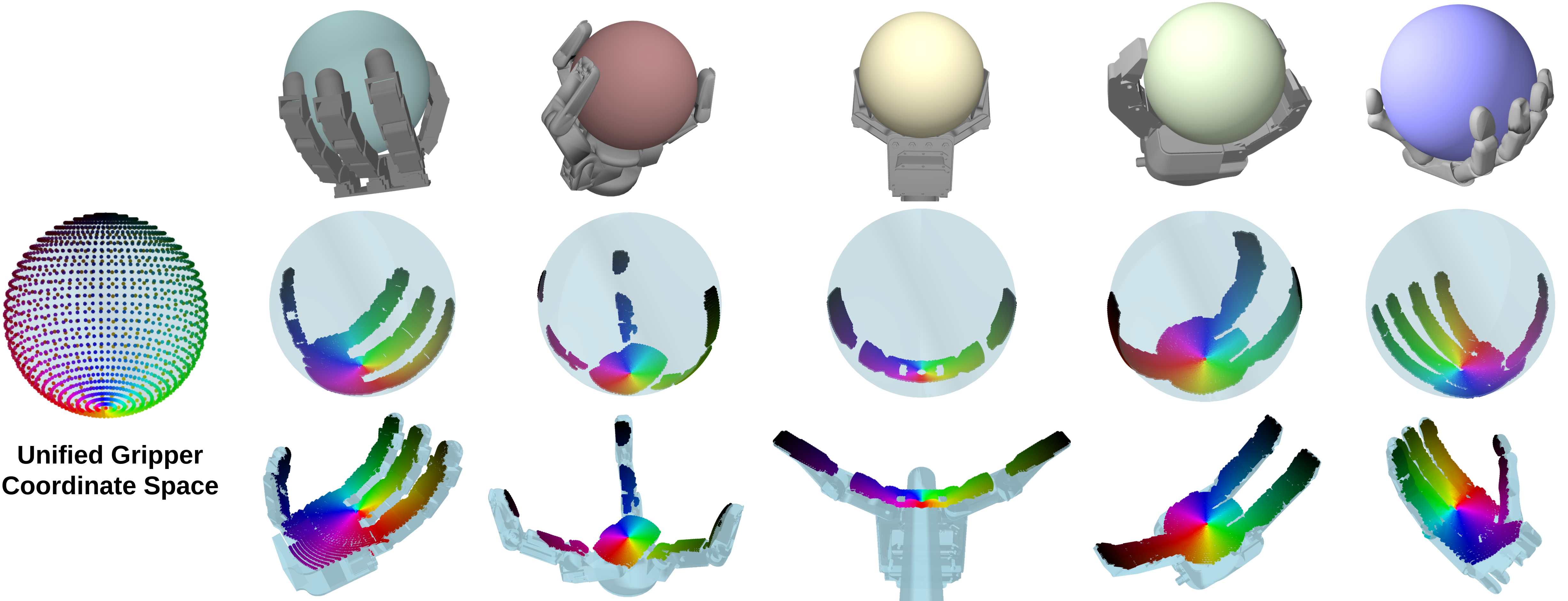}
    \caption{Visualization of Unified Gripper Coordinate System on grippers from MultiDex~\cite{li2023gendexgrasp} and their corresponding sphere points. The colors indicate where each gripper's region is mapped on the sphere surface.}
    \label{fig:uvcoords-viz-gripper}
    \vspace{-5mm}
    \end{center}
\end{figure*}
We use the maximal sphere for each gripper to assign a $(\lambda, \varphi)$ coordinate to the interior surface points of the gripper. Interior surface points are considered as they are close to the objects. This also reduces the ambiguity in which gripper points are supposed to contact the object, leading to stable grasps. The interior gripper surface points set $P_G$ is computed by uniformly sampling a set $R$ of unit ray directions from the sphere center and checking for ray-mesh intersection $\mathcal{I}_1(\hat{\mathbf{r}}, G)$ with the mesh $G$ of the gripper:
\begin{equation}\label{eq:rfp-gripper-surface-pts}
    P_G = \{ \mathbf{v}_g \in G \; | \; \exists \; \hat{\mathbf{r}} \in R, \; \text{s.t.,} \; \mathcal{I}_1(\hat{\mathbf{r}}, G) = \mathbf{v}_g      \}, 
\end{equation}
Here $\mathbf{v}_g$ is the intersection point of the ray $\hat{\mathbf{r}}$ and the gripper mesh $G$.
Each intersecting unit ray direction $\hat{\mathbf{r}}$ is also viewed as a surface point on a unit sphere with spherical coordinate $(\lambda, \varphi)$. 
We then assign the ray's angular component of the spherical coordinate as the gripper point's coordinate: \(
(\lambda_{\mathbf{v}_g}, \varphi_{\mathbf{v}_g}) := (\lambda_{\hat{\mathbf{r}}}, \varphi_{\hat{\mathbf{r}}})
\).
With a uniform set of rays, we obtain dense gripper surface points and their associated coordinates. For $M$ gripper points, we obtain the UGCS representation with shape $(M,2)$ and denote this as $\Phi_G$. Note that unlike manually selected sparse gripper key points, the UGCS representation is not affected by the ordering of points or the number $M$. Fig.~\ref{fig:uvcoords-viz-gripper} illustrates the idea behind obtaining $\Phi_G$ for each gripper. The gripper point cloud \(P_G \; ; |P_G| = M\) and its UGCS coordinate map \(\Phi_G\) are fixed for the gripper across its different grasps on objects. 


\begin{figure}[!ht]
    \centering
    \begin{center} 
    \includegraphics[width=0.7\linewidth]{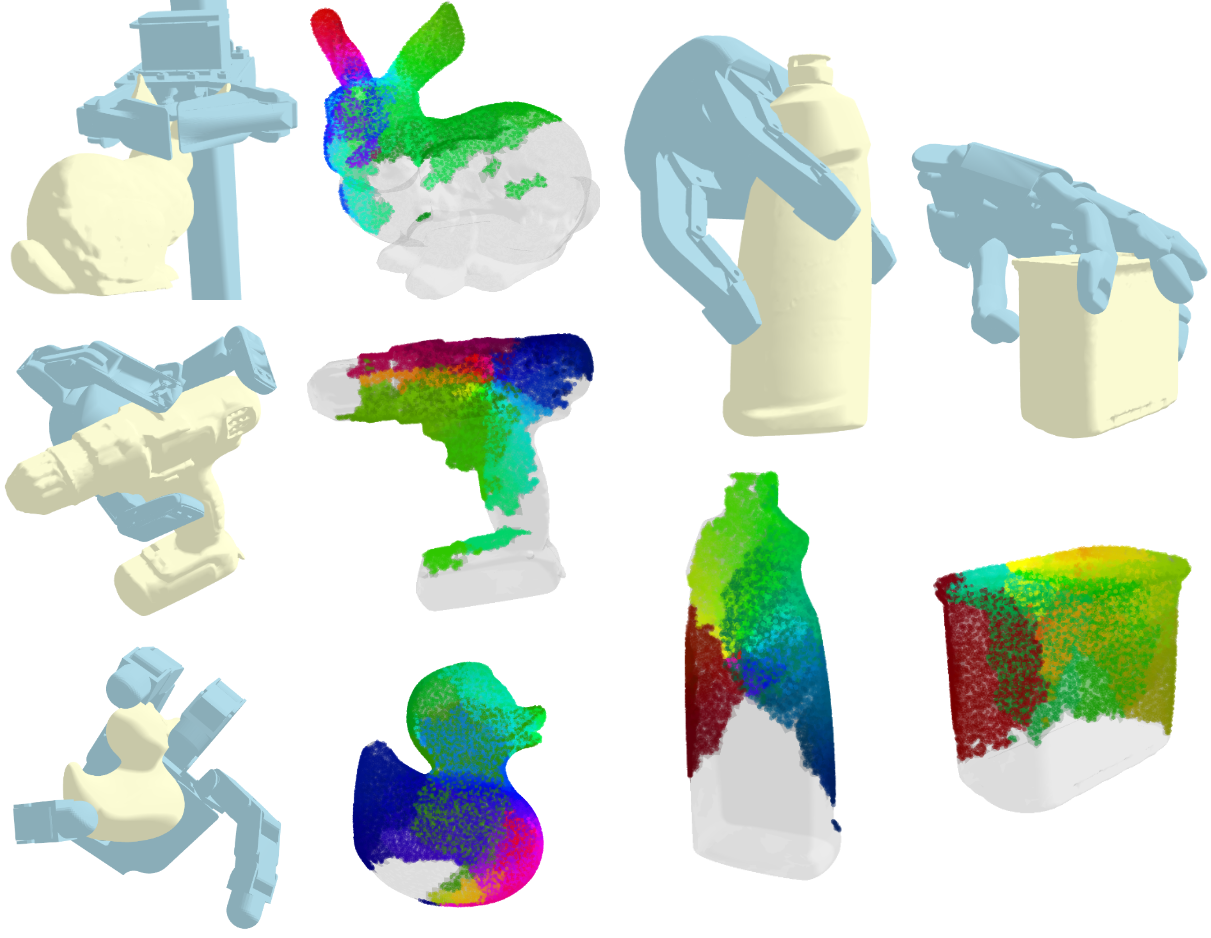}
    \caption{Ground-truth coordinate map $\Phi_O$ on the object set with some sample grasps from the MultiDex~\cite{li2023gendexgrasp} dataset.}
    \label{fig:uvcoords-viz-object}
    \vspace{-6mm}
    \end{center}
\end{figure}

\subsection{Grasp Representation}
In this section, we detail how to represent grasps in an object-centric manner using the UGCS coordinate map \(\Phi\) on the object's point cloud. 
Consider an object \(O\) 's point cloud $P_O$ and a grasp from gripper $G$ with its fixed surface point set \(P_G\) and coordinates $(\lambda, \varphi)$ for each gripper point.
To construct the coordinate map $\Phi_O$ of the object, we assign coordinates from the closest gripper point:
\begin{equation}\label{eq:rfp-object-coordinate-map}
\Phi_O^{(\mathbf{v}_o)}  =   \Phi_G^{(\mathbf{v}_g^*)}, \; \text{where} \; \mathbf{v}_g^{*}  =   \argmin_{\mathbf{v}_g \in P_G} || \mathbf{v}_o - \mathbf{v}_g ||,    
\end{equation}
Here $\mathbf{v}_o$ indicates an object point. In this way, we obtain the UGCS coordinate for each object point. The object map $\Phi_O$ provides an alternate grasp representation on the object, and its object-centric form is not constrained by the differences in grippers. For non-contact object points in $P_O$ (determined by a threshold of 1cm on gripper-object distance), we simply map them to a default coordinate of $(\lambda=0, \varphi=0)$. This is the sphere's north pole with original coordinates as $(-\pi/2, -\pi)$, and is never mapped to a gripper point as a consequence of maximal sphere test (see Fig.~\ref{fig:uvcoords-viz-gripper}). 
Finally, similar to a contact-map dataset on multi-gripper grasps, we create a coordinate-map dataset with grasps on the objects. Fig.~\ref{fig:uvcoords-viz-object} shows some examples of such grasp coordinate maps on object surface with the no-contact regions being transparent.

\section{Grasp Synthesis}\label{sec:method-grasp-synthesis}
\subsection{Model Architecture}
\begin{figure*}[h]
    \centering
    \begin{center} 
    \includegraphics[width=0.85\textwidth]{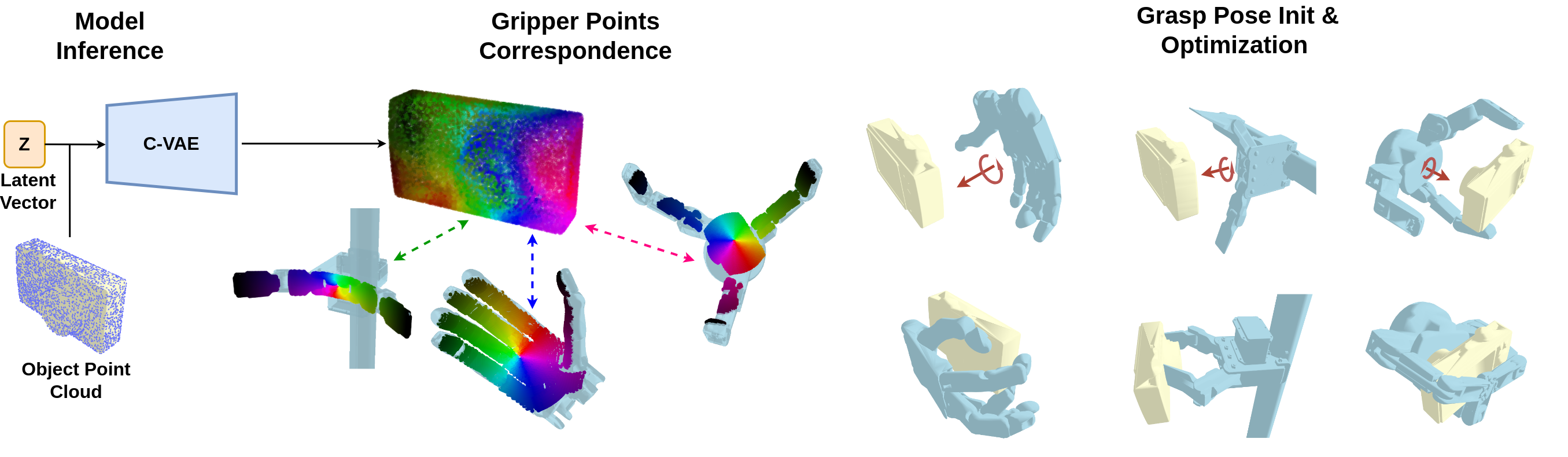}
    \caption{Method Overview: A randomly sampled latent vector is decoded to a probable object coordinate map $\hat{\Phi}_O$. Given a target (possibly unseen) gripper, correspondence is established between object and gripper points using $\hat{\Phi}_O$. $\hat{\Phi}_O$ also guides the pose initialization for the grasp optimization which results in successful grasps. }
    \label{fig:method-overview}
    \vspace{-6mm}
    \end{center}
\end{figure*}
One of the goals for RFP is to synthesize grasps on in/out of domain objects and grippers. In our modeling setup, we assume that we are given an object point cloud $P_O$ with $N$ points. The prediction problem consists of inferring an intermediate grasp representation of the object coordinate map $\Phi_O$ with values $(\lambda, \varphi)$ for $N$ points. We pose this as a generative modeling problem and utilize the CVAE architecture used in previous work on hand-object grasping~\cite{jiang2021hand, li2023gendexgrasp}.

The goal of the learning model is to reason about the grasps on the object $O$ and predict the grasp representations in the form of object coordinate maps $\Phi_O$ for each grasp on it.
Given $P_O$ and its coordinate map $\Phi_O$, the encoder $p_{\alpha}$ of the CVAE model learns to encode the object point cloud and the coordinate map into a latent vector $\mathbf{z}$ by modeling the distribution $p_{\alpha}(\mathbf{z} | P_O, \Phi_O )$.  The decoder $p_{\beta}$ learns to predict an accurate $\hat{\Phi}_O$ on the object given a randomly sampled $\mathbf{z}$  by modeling the distribution $p_{\beta}(\Phi_O | \mathbf{z}, P_O)$. We utilize a PointNet++~\cite{qi2017pointnet++} based backbone, where the final output consists of $(\hat{\lambda}, \hat{\varphi})$ coordinate prediction for each point in the object point cloud.

The model is trained using a standard loss formulation for CVAE with two components for the objective function. The first component is the reconstruction loss $L_{\text{recon}}$ on the ground truth coordinate map $\Phi$ and its prediction $\hat{\Phi}$: $L_{\text{recon}} = \sum_{i=1}^N || \Phi_O^{(i)} - \hat{\Phi}_O^{(i)}||$. The second component consists of a KL-divergence term $L_{\text{KL}}$ between a prior distribution on the latent space $p_{z} = \mathcal{N}(0, \mathcal{I})$ and the approximate posterior modeled by the decoder $p_{\beta}(\Phi_O | \mathbf{z}, P_O) $: $L_{\text{KL}} = D_{\text{KL}} (p_{z}, p_{\beta})$. The total loss is a weighted sum between the two: \( L_{\text{total}} = \; L_{\text{recon}} \; + \; w_{\text{KL}} \; L_{\text{KL}} \), where the weight \(w_{\text{KL}}\) is annealed over training epochs. During inference, given an object point cloud, we sample a latent vector $\mathbf{z} \sim \mathcal{N}(0,\mathcal{I})$ and pass it through the decoder with the object point cloud to obtain the predicted coordinate map $\hat{\Phi}_O$. This output is then used as an intermediate representation for grasp optimization on a given gripper as shown in Fig.~\ref{fig:method-overview}.

\subsection{Grasp Optimization}
\subsubsection{Coordinate Map Correspondence}
The final output for grasp synthesis should be a grasp configuration $\mathbf{q}_G$, with gripper pose and joint values. We generate $\mathbf{q}_G$ for a predicted object $O$'s coordinate map $\hat{\Phi}_O$ and target gripper $G$ via a UGCS correspondence optimization. 
In contact map based grasp synthesis, predicted contact map is oblivious to the gripper fingers and hence ambiguous to interpret. A key advantage of our method is that we can easily establish a correspondence between object and gripper points via the coordinate maps.
Given $N$ object points, and $M$ gripper points, correspondence is found by assigning each object point $\mathbf{v}_o$ to a gripper point $\mathbf{v}_g^{*(o)}$. Assignment is based on the great-circle distance $D_H$ (Haversine distance) between their respective coordinates $(\hat{\lambda}_{\mathbf{v}_o}, \hat{\varphi}_{\mathbf{v}_o})$ and $(\lambda_{\mathbf{v}_g}, \varphi_{\mathbf{v}_g})$:
\begin{equation}\label{eq:rfp-coordinate-correspondence}
\mathbf{v}_{g}^{*(o)} = \argmin_{\mathbf{v}_g \in P_G} \; \; D_H  \left( (\hat{\lambda}_{\mathbf{v}_o}, \hat{\varphi}_{\mathbf{v}_o})  \; ; \; (\lambda_{\mathbf{v}_g}, \varphi_{\mathbf{v}_g})   \right).
\end{equation}
This establishes a dense mapping with two matching subsets of $P_{O_c}$ and $P_{G_c}$ from the object and gripper points.


\subsubsection{Pose Initialization}
Randomly sampling gripper poses around the object for initialization like in prior works can lead to suboptimal grasps as the optimization gets stuck in local minima. 
In addition to the correspondence, the predicted coordinate map $\hat{\Phi}_O$ contains another signal to the optimization in terms of gripper pose initialization. By definition, in UGCS, each gripper grasps its sphere with palm/base touching the sphere's south pole with a normalized coordinate value $(\lambda=0, \varphi=1)$. Using this fact, we infer a set $S_B$ of probable object points that map to the gripper base/palm region. $S_B$ contains the points whose
predicted coordinates $(\hat{\lambda}, \hat{\varphi}) \in \hat{\Phi}_O$ are within some thresholds $\lambda_{\text{ub}}$ and $\varphi_{\text{lb}}$ for base pole's coordinates:
\begin{equation} \label{eqn:rfp-pose-init}  
    S_B = \{ \mathbf{v}_o \in P_O \; s.t., \;  \hat{\lambda}_{\mathbf{v}_o} \leq \lambda_{\text{ub}} \;\; \text{and} \;\; \hat{\varphi}_{\mathbf{v}_o} \geq \varphi_{\text{lb}}  \}.
\end{equation}
The initial gripper position is set to $0.1m$ away from the mean location along the mean surface normal for palm-matching point set $S_B$. The approach direction for the gripper palm is opposite to the surface normal as shown in Fig.~\ref{fig:method-overview}. 

\begin{table*}[!ht]
\centering
\caption{Comparison of RFP with prior works on success rate and diversity metrics}
\label{tab:full-comparison}
\resizebox{0.75\linewidth}{!}{
    \begin{tabular}{c|ccc|c|ccc|c}
\toprule
\multicolumn{1}{c|}{\multirow{2}{*}{Method}} & \multicolumn{4}{c|}{Success Rate (\%) ↑} & \multicolumn{4}{c}{Diversity (rad) ↑}  \\ 
\multicolumn{1}{c|}{}                        & \multicolumn{1}{c|}{ezgripper} & \multicolumn{1}{c|}{barrett} & \multicolumn{1}{c|}{shadowhand} & \multicolumn{1}{c|}{\textbf{Mean}} & \multicolumn{1}{c|}{ezgripper} & \multicolumn{1}{c|}{barrett} & \multicolumn{1}{c|}{shadowhand} & \multicolumn{1}{c}{\textbf{Mean}} \\ \midrule 
DFC \cite{liu2021synthesizing}                          & 58.81                          & 85.48                        & 72.86                           & 72.38                              & \textbf{0.3095}                & 0.3770                       & \textbf{0.3472}                 & \textbf{0.3446}                    \\
AdaGrasp \cite{xu2021adagrasp}                          & 60.00                          & 80.00                        & -                               & 70.00                              & 0.0003                         & 0.0002                       & -                               & 0.0003                             \\
GenDexGrasp \cite{li2023gendexgrasp}                          & 43.44                          & 71.72                        & \textbf{77.03}                  & 64.01                              & 0.2380                         & 0.2480                       & 0.2110                          & 0.2323                             \\
GeoMatch \cite{geomatch}                               & \textbf{75.00}                 & \textbf{90.00}               & 72.50                           & 79.17                              & 0.1880                         & 0.2490                       & 0.2050                          & 0.2140                             \\
\midrule \textbf{RFP (Ours)}                           & 72.34                          & \textbf{90.00}               & 76.87                           & \textbf{79.73}                     & 0.2330                         & \textbf{0.4240}              & 0.2520                          & 0.3030 \\
 \textbf{RFP* (w/o refinement)}            & 52.5                           & 86.56                        & 55.47                           & 64.84                              & 0.2326                         & 0.4256                        & 0.2519                         &  0.3034     \\
\bottomrule
\end{tabular}
}
\vspace{-3mm}
\end{table*}

\subsubsection{Optimization}\label{sec:method-graspopt}
The grasp configuration $\mathbf{q}_G$ is computed by minimizing an objective function that considers hand-object penetration, joint value validity, and hand-object distance. A differentiable kinematics gripper model provides gradients on $\mathbf{q}$. Here, the hand-object distance \(E_{\text{dist}}\) is computed only with corresponding point pairs from $P_{O_c}$ and $P_{G_c}$. 
\begin{equation}\label{eq:rfp-grasp-opt}
\mathbf{q}_G^* = \argmin_{\mathbf{q}} E_{\text{dist}}(P_{O_c}, P_{G_c} ; \mathbf{q}) + E_{\text{p}}({O}, G ; \mathbf{q})  + E_{\text{n}}(\mathbf{q}).
\end{equation}
The terms for hand-object collision $E_{\text{p}}$ and gripper joint validity $E_\text{n}$ are adopted from~\cite{li2023gendexgrasp}. They constrain the grasp $\mathbf{q}$ to avoid collisions, and keep the joint values within limits. The correspondence-driven $E_{\text{dist}}$ acts as a denser version of inverse kinematics with sparse keypoints seen in prior works.

\begin{figure}[!ht]
    \centering
    \begin{center} 
        \includegraphics[width=0.95\linewidth]{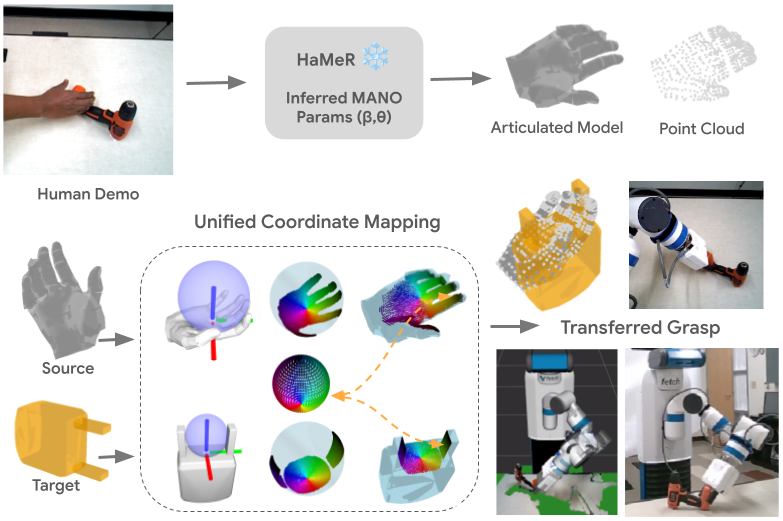}
        \caption{Grasp Transfer Pipeline: Human demonstration to executing transferred grasp on Fetch}
        \label{fig:human-grasp-transfer}
        \vspace{-6mm}
        \end{center}
\end{figure}

\section{Grasp Transfer}
\label{sec:method-rfp-grasp-transfer}
The unified gripper coordinate space also enables grasp transfer between any two (potentially unseen) grippers without retargeting. Any new, unseen gripper is fit to UGCS via its maximal sphere and subsequently obtaining its surface point coordinate map.
Suppose we are given two grippers, \(s\) (source) and \(t\) (target) with their coordinate maps \(\Phi_{s}, \Phi_{t}\). 
In the grasp transfer problem, we are given a grasp \(\mathbf{q}_s\) for the source and we need to infer a similar grasp \(\mathbf{q}_t\) for the target. \(\mathbf{q}_t\) is inferred via the grasp transfer optimization shown in Eq.~\eqref{eq:rfp-grasp-transfer}. 
Here we do not assume access to object point cloud for the source gripper demonstration and consequently, do not involve hand-object point correspondence and collision terms.
The two grasps \(\mathbf{q}_s, \mathbf{q}_t\) give us the transformed gripper point clouds \(P_s, P_t\) by applying the grasps to respective grippers. A set of mutually corresponding pairs of points \(P_{s}^{(c)}, P_{t}^{(c)}\) via their respective coordinate maps \(\Phi_{t}, \Phi_{s}\) is computed using a similar formulation seen in Eq.~\eqref{eq:rfp-coordinate-correspondence}. Given the corresponding set, we enforce a distance similarity between the point sets so that the target gripper points align with the source gripper. 
\begin{equation}\label{eq:rfp-grasp-transfer}
\mathbf{q}_t^* = \argmin_{\mathbf{q}} \;  E_{\text{dist}}(P_{s}^{(c)}, P_{t}^{(c)} ; \mathbf{q}) + E_{\text{n}}(\mathbf{q}),   
\end{equation}
where $E_{\text{dist}}$ and $E_{\text{n}}$ are defined as in Eq.~\eqref{eq:rfp-grasp-opt}. Note that going from dexterous grippers to simple, 2-finger gripper is not well defined for plain contact-map driven transfer. The dense correspondences from our method resolve this ambiguity and output a functional and similar grasping pose.
An illustration of the grasp transfer pipeline between human hand and Fetch robot gripper is shown in Fig.~\ref{fig:human-grasp-transfer}.

%% file: includes/sec4-exp.tex
\section{Experiments}
\label{sec:experiments}

\subsection{Grasp Synthesis in Simulation}
We used the MultiDex dataset~\cite{li2023gendexgrasp} to train and evaluate our models in simulation making it comparable to prior works~\cite{li2023gendexgrasp, geomatch} and their evaluation protocol. 
Each grasp generated by our method is tested for 6 seconds with 60 simulation steps per second and by applying a consistent acceleration of $0.5ms^{-2}$ on the object in the NVIDIA Isaac Gym simulation. 
A grasp is considered as failure if the object moves by 2cm at the end of any test else it is labeled as successful.
We record the same evaluation metrics of grasp success rate and the diversity in successful grasps (standard deviation of joint values) in order to be consistent with the comparison. The test set was comprised of the same split of 10 unseen objects as in~\cite{li2023gendexgrasp, geomatch}, and we sampled 64 grasps for each gripper-object pair for evaluation.

\subsubsection{Grasp Synthesis with Seen Grippers}

\begin{figure}[h]
\centering
\begin{center} 
\includegraphics[width=0.7\linewidth]{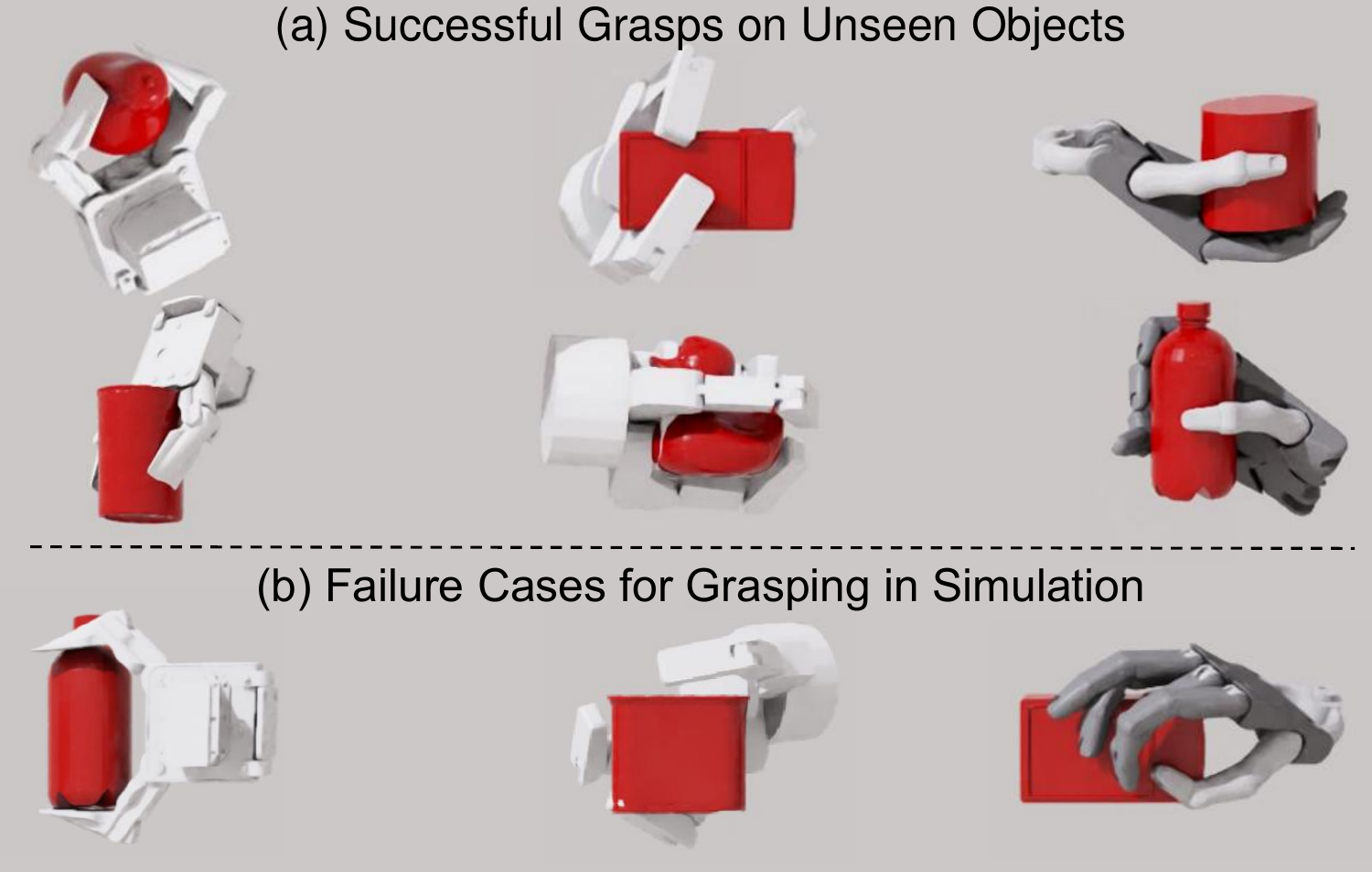}
\caption{Simulation Results: (a) Successful grasps on unseen objects with RFP ; (b) Some failure modes: insufficient force closure or not using all fingers}
\label{fig:results-success_fail_grasps}
\end{center}
\vspace{-6mm}
\end{figure}

We evaluated the RFP by training with all 5 grippers and evaluating on the test object set. We report metrics for Ezgripper, Barrett and Shadowhand according to~\cite{li2023gendexgrasp, geomatch}. Table~\ref{tab:full-comparison} shows the success rate and diversity of all the methods using models trained on all the grippers. Results for DFC~\cite{liu2021synthesizing}, AdaGrasp~\cite{xu2021adagrasp}, GenDexGrasp~\cite{li2023gendexgrasp} and GeoMatch~\cite{geomatch} are taken from GeoMatch~\cite{geomatch}. We observe that RFP outperforms comparable methods in terms of grasp diversity and success rate. Our results are comparable to~\cite{geomatch} and better than~\cite{li2023gendexgrasp}. It is important to note that while the RFP demonstrates competitive performance with~\cite{geomatch}, it is capable of dealing with \textit{unseen grippers} whereas~\cite{geomatch} cannot do so. Fig.~\ref{fig:results-success_fail_grasps} shows some successful and failure grasps of our method.

\begin{table}
\centering
\caption{Performance on unseen grippers for RobotFingerPrint (RFP) and GenDexGrasp (GDG)~\cite{li2023gendexgrasp}. RFP* represents the experiment with no-refinement ablation.}
\label{tab:comparison_out_domain_gendexgrasp}
\resizebox{0.8\linewidth}{!}{
\begin{tabular}{c|c|cc}
\toprule
\multicolumn{1}{c|}{Gripper} & \multicolumn{1}{c|}{Method} & \multicolumn{1}{c}{Succ. Rate (\%) ↑} & \multicolumn{1}{c}{Diversity (rad) ↑} \\ \midrule

\multirow{2}{*}{Ezgripper} & RFP  & \textbf{63.28} & 0.224 \\
                           & RFP* & 42.97 & 0.214 \\ 
                           & GDG~\cite{li2023gendexgrasp}  & 38.59 & \textbf{0.248} \\\midrule

\multirow{2}{*}{Barrett}   & RFP  & \textbf{83.90} & \textbf{0.400} \\
                           & RFP* & 77.96 & 0.396 \\
                           & GDG~\cite{li2023gendexgrasp}  & 70.31 & 0.267 \\\midrule

\multirow{2}{*}{Shadow}    & RFP  & 67.96 & \textbf{ 0.238} \\
                           & RFP* & 51.09 & 0.238 \\
                           & GDG~\cite{li2023gendexgrasp}  & \textbf{77.19} & 0.207 \\
  \bottomrule
\end{tabular}
}
\vspace{-4mm}
\end{table}

\subsubsection{Grasp Synthesis with Unseen Grippers}
Furthermore, we compared the performance of our method with that of GenDexGrasp~\cite{li2023gendexgrasp} when using out-of-domain grippers and the same test set of unseen objects. The model for an out-of-domain gripper was trained with grasps from all grippers except itself. The comparison is presented in Table~\ref{tab:comparison_out_domain_gendexgrasp} where we see a strong improvement over the prior work across most of the metrics. RFP under-performs GenDexGrasp with ShadowHand in success rate but shows better diversity. It could be due to difficulty in setting up the correspondence for such a dexterous gripper in presence of noisy coordinate predictions. On the other hand, RFP outperforms in the case of Ezgripper and Barrett as the correspondence helps the coordinate mapping. With GenDexGrasp, the multi-fingered contact maps from other grippers could induce uncertainty for the finger mapping with Ezgripper or Barrett.

\begin{figure}[!ht]
    \centering
    \begin{center} 
    \includegraphics[width=0.7\linewidth]{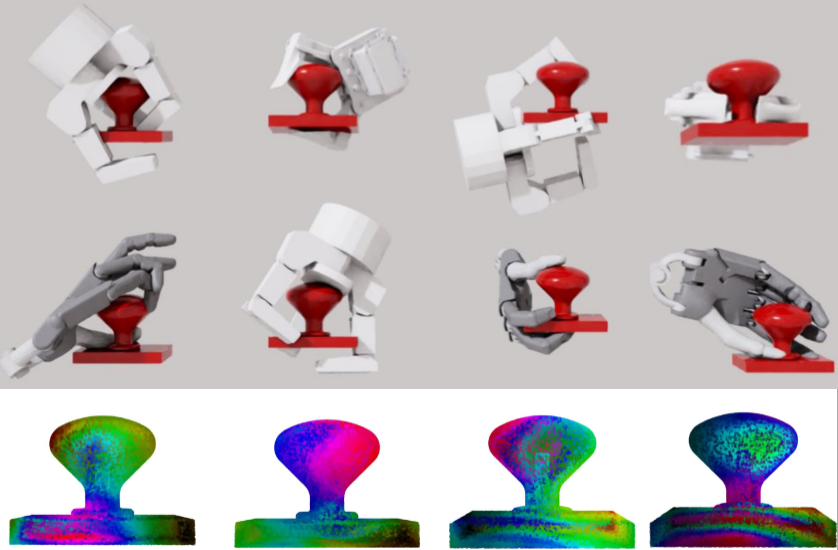}
    \caption{Similar grasp prediction across different grippers using the same predicted object coordinate map $\hat{\Phi}_O$.}
    \label{fig:uvmap_grasps_isaac_sim}
    \vspace{-6mm}
    \end{center}
\end{figure}


\begin{figure*}
    \centering
    \begin{center} 
        \includegraphics[width=0.8\linewidth]{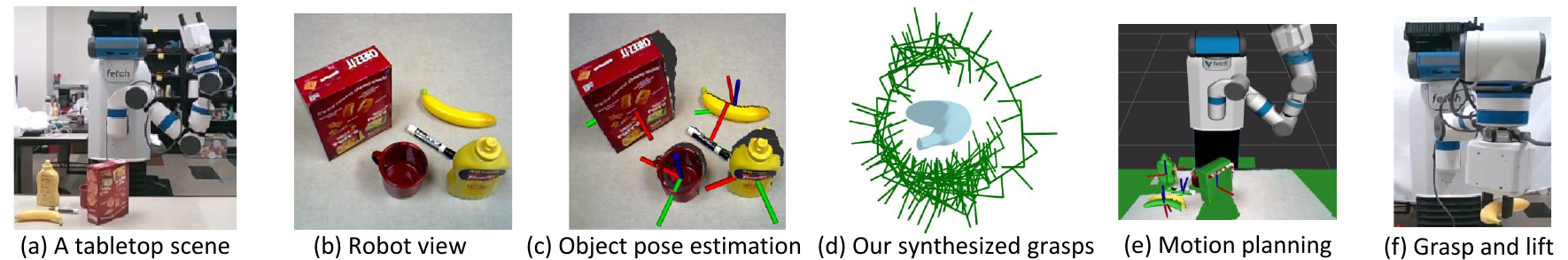}
        \caption{Illustration of our real-world grasping pipeline. Our method is used to synthesize grasps given a 3D object model.}
        \label{fig:scene_replica}
        \vspace{-4mm}
        \end{center}
\end{figure*}

\begin{figure*}
    \centering
    \begin{center} 
        \includegraphics[width=0.75\linewidth]{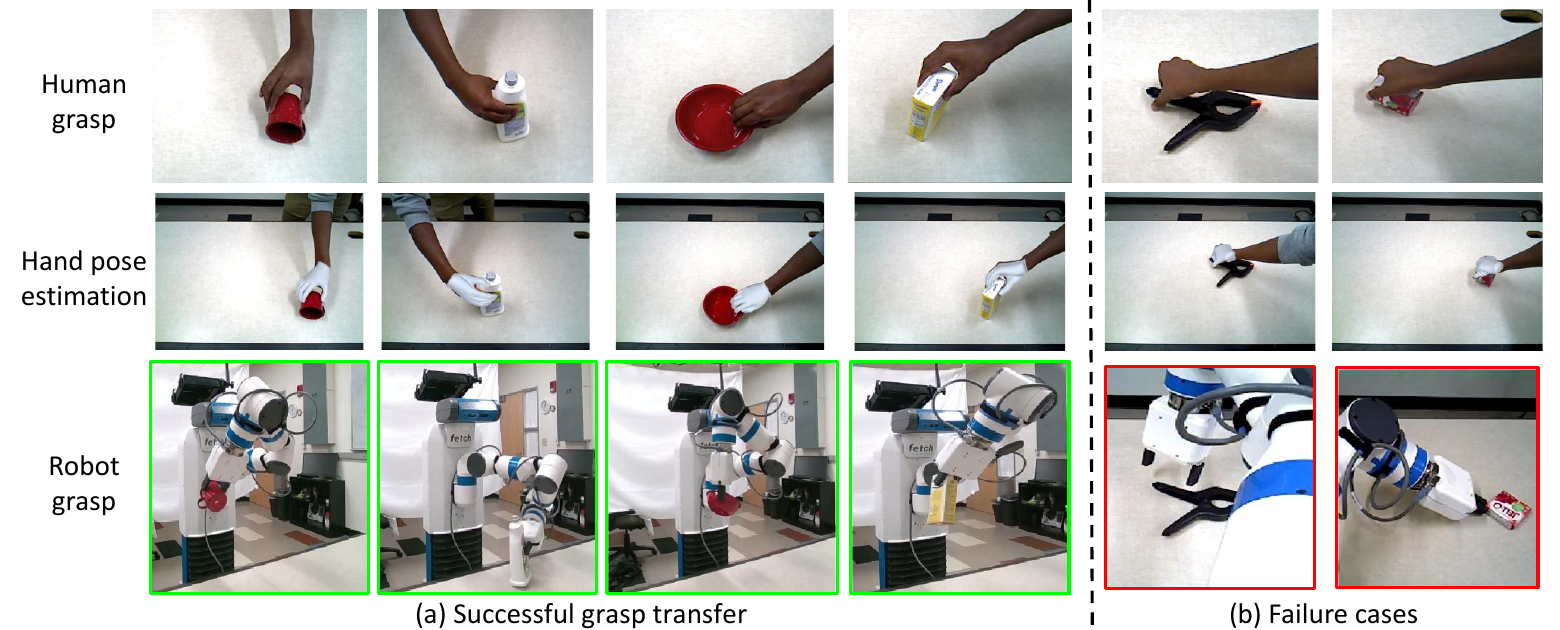}
        \caption{Grasp Transfer Experiments: Examples of huamn demo (from robot's view) with success/failure cases.}
        \label{fig:exp-grasp-transfer}
        \vspace{-6mm}
        \end{center}
\end{figure*}

\subsection{Ablation Studies}
\subsubsection{Grasp Refinement during Simulation Test}
We carried out an ablation study for the grasp refinement step used in the simulation evaluation tests as in~\cite{li2023gendexgrasp, geomatch}. A contact-aware refinement is applied to the predicted grasp $\mathbf{q}$ via a single gradient update step to obtain force closure around the object. By removing this refinement, we wanted to test the efficacy of our method in terms of its original output. We compare it with the refined outputs from other methods under both in-domain and out-domain settings for the grippers. As seen in Tables~\ref{tab:full-comparison} and Table~\ref{tab:comparison_out_domain_gendexgrasp}, RFP without refinement remains competitive even under the removal of this refinement step and establishes strong baselines. It even outperforms on some metrics from prior works~\cite{li2023gendexgrasp, liu2021synthesizing, xu2021adagrasp} which include this refinement during their evaluation. The dense correspondence-driven grasp optimization and a stable pose initialization are able to help the optimization, avoiding bad minima and converging to good quality grasps.   

\subsubsection{Grasp Inference on Same Object Coordinate Map}
On the same predicted coordinate map of the object, we observe that the three different grippers grasp the object in a similar fashion, especially in terms of approach and position around the object as visualized in Fig.~\ref{fig:uvmap_grasps_isaac_sim}. The mean difference in grasp positions was $0.03m$ for Barrett and Ezgripper while it was circa $0.06m$ for both of them with the Shadowhand grasps. This indicates the efficacy of RFP in generalizing the grasps from different grippers in a unified coordinate space. 


\subsection{Real World Grasping Evaluation}
We used the SceneReplica \cite{khargonkar2024scenereplica} real-world benchmark to evaluate the grasps generated by our method. SceneReplica benchmark is comprised of 20 standardized cluttered scenes with 5 YCB objects each. It benchmarks the task of grasping each object in scene and moving to a dropoff location with 100 trials in total. Grasping and pick-place success rates were the evaluation metrics used for the trials. An example setup of the experiments is shown in Fig.~\ref{fig:scene_replica}. 
For our model-based pipeline, we utilized the GDRNPP model~\cite{liu2022gdrnpp_bop} for object pose estimation from the robot’s RGB-D camera. Grasp generation was performed for each object using RFP by inferring coordinate maps on the object using the trained model, and optimizing for the Fetch gripper's grasp configuration. The MultiGripperGrasp~\cite{casas2024multigrippergrasp} toolkit was used to rank the grasps based on a fall time evaluation metric in simulation. Moveit was used for motion planning to evaluate and execute collision-free grasps in descending order of evaluation scores. 

It is worth noting that of these 16 YCB objects, 6 objects for which we are generating grasps are unseen to our model. Moreover, the Fetch robot's 2 finger gripper is also a new gripper outside of the model's training domain. The grasp was deemed successful if the robot lifted the object, with the pick-and-place task failing only if the object fell before placement. As seen in Table~\ref{tab:rfp-scene-replica}, the grasps computed by RFP outperform those given by GraspIt! on the benchmark. Despite the training dataset not containing grasps from an equivalent 2-finger gripper, RFP is able to synthesize reasonable real-world grasps for the Fetch gripper. Please see supplementary video for grasping examples.

\begin{table}
\centering
\caption{Grasping and Pick-and-Place (P\&P) success rates on the SceneReplica benchmark \cite{khargonkar2024scenereplica}  for 100 trials.}
\label{tab:rfp-scene-replica}
\resizebox{0.8\linewidth}{!}{
\begin{tabular}{c|c|c|c}
\toprule
\multicolumn{1}{c|}{Perception Module} & \multicolumn{1}{c|}{Grasp Module} & \multicolumn{1}{c}{P\&P (\%) ↑} & \multicolumn{1}{c}{Grasping (\%)↑} \\ \midrule
GDRNPP \cite{liu2022gdrnpp_bop} & RFP (\textbf{ours}) & \textbf{70} & \textbf{73} \\ \midrule
GDRNPP \cite{liu2022gdrnpp_bop} & GraspIt! \cite{miller2004graspit}  & 66 & 69 \\\midrule
PoseRBPF \cite{deng2019pose} & GraspIt! \cite{miller2004graspit} & 58 & 64 \\ 
  \bottomrule
\end{tabular}
}
\vspace{-4mm}
\end{table}

\subsection{Human-to-Robot Grasp Transfer}
A key advantage of the RFP over previous works is the ability to fit novel and unseen grippers at test time via the unified coordinate map, e.g. from a human hand model.  
We evaluated grasp transfer from humans by collecting human grasp demonstrations and executing the transferred grasps with a Fetch robot gripper, shown in Fig.~\ref{fig:human-grasp-transfer}. For a human demonstration, we infer the MANO~\cite{romero2017embodied} hand parameters with the HaMeR~\cite{pavlakos2024reconstructing} model. 
Using the sphere coordinate mapping defined in Sec.~\ref{sec:method-ugcs-gripper-pts}, we have the correspondences for human model to the Fetch gripper used for grasp transfer. We evaluated the grasp transfer pipeline in real world on 14 YCB objects from SceneReplica~\cite{khargonkar2024scenereplica} by making the Fetch robot grasp and lift the object using the transfer method. 
We achieve a grasping success rate of \(0.785\) ($11/14$), with examples shown in Fig.~\ref{fig:exp-grasp-transfer}. Most failures stem from an incorrect initialization from noisy human hand pose inference.

%% file: includes/sec5-conclusion.tex
\section{Conclusion and Future Work}
\label{sec:conclusion}
We introduce RobotFingerPrint (RFP), a novel and effective way to unify grasp representations across different grippers into a common coordinate space. The unified grasp representation requires no manual specification and has a precise, dense mapping for each gripper's individual fingers. This helps in synthesizing diverse and stable grasp configurations via (1) hand-object correspondence and (2) well-initialized grasp pose. RFP shows strong performance across several evaluation scenarios in both simulation and real-world testing with unseen Fetch gripper. Further, it enables direct grasp transfer between novel grippers without manual labeling and enables downstream applications in learning from human demonstrations. Future work would be to address a potential limitation in collision-aware grasp generation on partial object observations, and real-time grasp inference for use in a closed-loop setting.



%% file: root.bbl
\begin{thebibliography}{10}
\providecommand{\url}[1]{#1}
\csname url@samestyle\endcsname
\providecommand{\newblock}{\relax}
\providecommand{\bibinfo}[2]{#2}
\providecommand{\BIBentrySTDinterwordspacing}{\spaceskip=0pt\relax}
\providecommand{\BIBentryALTinterwordstretchfactor}{4}
\providecommand{\BIBentryALTinterwordspacing}{\spaceskip=\fontdimen2\font plus
\BIBentryALTinterwordstretchfactor\fontdimen3\font minus \fontdimen4\font\relax}
\providecommand{\BIBforeignlanguage}[2]{{%
\expandafter\ifx\csname l@#1\endcsname\relax
\typeout{** WARNING: IEEEtran.bst: No hyphenation pattern has been}%
\typeout{** loaded for the language `#1'. Using the pattern for}%
\typeout{** the default language instead.}%
\else
\language=\csname l@#1\endcsname
\fi
#2}}
\providecommand{\BIBdecl}{\relax}
\BIBdecl

\bibitem{miller2004graspit}
A.~T. Miller and P.~K. Allen, ``Graspit! a versatile simulator for robotic grasping,'' \emph{IEEE Robotics \& Automation Magazine}, 2004.

\bibitem{diankov2008openrave}
R.~Diankov and J.~Kuffner, ``{OpenRAVE}: A planning architecture for autonomous robotics,'' \emph{Robotics Institute, Pittsburgh, PA, Tech. Rep. CMU-RI-TR-08-34}, vol.~79, 2008.

\bibitem{nguyen1988constructing}
V.-D. Nguyen, ``Constructing force-closure grasps,'' \emph{The International Journal of Robotics Research}, vol.~7, no.~3, pp. 3--16, 1988.

\bibitem{mousavian20196}
A.~Mousavian, C.~Eppner, and D.~Fox, ``6-dof graspnet: Variational grasp generation for object manipulation,'' in \emph{Proceedings of the IEEE/CVF international conference on computer vision}, 2019.

\bibitem{li2023gendexgrasp}
P.~Li, T.~Liu, Y.~Li, Y.~Geng, Y.~Zhu, Y.~Yang, and S.~Huang, ``Gendexgrasp: Generalizable dexterous grasping,'' in \emph{2023 IEEE International Conference on Robotics and Automation (ICRA)}, 2023.

\bibitem{urain2023se}
J.~Urain, N.~Funk, J.~Peters, and G.~Chalvatzaki, ``Se (3)-diffusionfields: Learning smooth cost functions for joint grasp and motion optimization through diffusion,'' in \emph{2023 IEEE International Conference on Robotics and Automation (ICRA)}, 2023.

\bibitem{eppner2021acronym}
C.~Eppner, A.~Mousavian, and D.~Fox, ``Acronym: A large-scale grasp dataset based on simulation,'' in \emph{IEEE International Conference on Robotics and Automation (ICRA)}, 2021, pp. 6222--6227.

\bibitem{turpin2023fast-graspd}
D.~Turpin, T.~Zhong, S.~Zhang, G.~Zhu, E.~Heiden, M.~Macklin, S.~Tsogkas, S.~Dickinson, and A.~Garg, ``Fast-grasp'd: Dexterous multi-finger grasp generation through differentiable simulation,'' 2023.

\bibitem{sundermeyer2021contact}
M.~Sundermeyer, A.~Mousavian, R.~Triebel, and D.~Fox, ``Contact-graspnet: Efficient 6-dof grasp generation in cluttered scenes,'' in \emph{2021 IEEE International Conference on Robotics and Automation (ICRA)}.\hskip 1em plus 0.5em minus 0.4em\relax IEEE, 2021, pp. 13\,438--13\,444.

\bibitem{mayer2022ffhnet}
V.~Mayer, Q.~Feng, J.~Deng, Y.~Shi, Z.~Chen, and A.~Knoll, ``Ffhnet: Generating multi-fingered robotic grasps for unknown objects in real-time,'' in \emph{2022 International Conference on Robotics and Automation (ICRA)}.\hskip 1em plus 0.5em minus 0.4em\relax IEEE, 2022, pp. 762--769.

\bibitem{shao2020unigrasp}
L.~Shao, F.~Ferreira, M.~Jorda, V.~Nambiar, J.~Luo, E.~Solowjow, J.~A. Ojea, O.~Khatib, and J.~Bohg, ``Unigrasp: Learning a unified model to grasp with multifingered robotic hands,'' \emph{IEEE Robotics and Automation Letters}, vol.~5, no.~2, pp. 2286--2293, 2020.

\bibitem{geomatch}
M.~Attarian, M.~A. Asif, J.~Liu, R.~Hari, A.~Garg, I.~Gilitschenski, and J.~Tompson, ``Geometry matching for multi-embodiment grasping,'' in \emph{Proceedings of the 7th Conference on Robot Learning (CoRL)}, 2023.

\bibitem{wang2019normalized}
H.~Wang, S.~Sridhar, J.~Huang, J.~Valentin, S.~Song, and L.~J. Guibas, ``Normalized object coordinate space for category-level 6d object pose and size estimation,'' in \emph{Proceedings of the IEEE/CVF Conference on Computer Vision and Pattern Recognition}, 2019, pp. 2642--2651.

\bibitem{guler2018densepose}
R.~A. G{\"u}ler, N.~Neverova, and I.~Kokkinos, ``Densepose: Dense human pose estimation in the wild,'' in \emph{Proceedings of the IEEE Conference on Computer Vision and Pattern Recognition}, 2018, pp. 7297--7306.

\bibitem{wise2016fetch}
M.~Wise, M.~Ferguson, D.~King, E.~Diehr, and D.~Dymesich, ``Fetch and freight: Standard platforms for service robot applications,'' in \emph{Workshop on autonomous mobile service robots}, 2016, pp. 1--6.

\bibitem{khargonkar2024scenereplica}
N.~Khargonkar, S.~H. Allu, Y.~Lu, B.~Prabhakaran, Y.~Xiang \emph{et~al.}, ``Scenereplica: Benchmarking real-world robot manipulation by creating replicable scenes,'' in \emph{2024 IEEE International Conference on Robotics and Automation (ICRA)}.\hskip 1em plus 0.5em minus 0.4em\relax IEEE, 2024, pp. 8258--8264.

\bibitem{ferrari1992planning}
C.~Ferrari and J.~F. Canny, ``Planning optimal grasps.'' in \emph{ICRA}, vol.~3, no.~4, 1992, p.~6.

\bibitem{borst2004grasp}
C.~Borst, M.~Fischer, and G.~Hirzinger, ``Grasp planning: How to choose a suitable task wrench space,'' in \emph{IEEE International Conference on Robotics and Automation, 2004. Proceedings. ICRA'04. 2004}, vol.~1.\hskip 1em plus 0.5em minus 0.4em\relax IEEE, 2004, pp. 319--325.

\bibitem{berenson2008grasp}
D.~Berenson and S.~S. Srinivasa, ``Grasp synthesis in cluttered environments for dexterous hands,'' in \emph{Humanoids 2008-8th IEEE-RAS International Conference on Humanoid Robots}.\hskip 1em plus 0.5em minus 0.4em\relax IEEE, 2008.

\bibitem{liu2021synthesizing}
T.~Liu, Z.~Liu, Z.~Jiao, Y.~Zhu, and S.-C. Zhu, ``Synthesizing diverse and physically stable grasps with arbitrary hand structures using differentiable force closure estimator,'' \emph{IEEE Robotics and Automation Letters}, vol.~7, no.~1, pp. 470--477, 2021.

\bibitem{wang2023dexgraspnet}
R.~Wang, J.~Zhang, J.~Chen, Y.~Xu, P.~Li, T.~Liu, and H.~Wang, ``Dexgraspnet: A large-scale robotic dexterous grasp dataset for general objects based on simulation,'' in \emph{2023 IEEE International Conference on Robotics and Automation (ICRA)}.\hskip 1em plus 0.5em minus 0.4em\relax IEEE, 2023, pp. 11\,359--11\,366.

\bibitem{casas2024multigrippergrasp}
L.~F. Casas, N.~Khargonkar, B.~Prabhakaran, and Y.~Xiang, ``Multigrippergrasp: A dataset for robotic grasping from parallel jaw grippers to dexterous hands,'' in \emph{2024 IEEE/RSJ International Conference on Intelligent Robots and Systems (IROS)}.\hskip 1em plus 0.5em minus 0.4em\relax IEEE, 2024, pp. 2978--2984.

\bibitem{fang2020graspnet1billion}
H.-S. Fang, C.~Wang, M.~Gou, and C.~Lu, ``Graspnet-1billion: A large-scale benchmark for general object grasping,'' in \emph{Proceedings of the IEEE/CVF conference on computer vision and pattern recognition}, 2020, pp. 11\,444--11\,453.

\bibitem{xu2021adagrasp}
Z.~Xu, B.~Qi, S.~Agrawal, and S.~Song, ``Adagrasp: Learning an adaptive gripper-aware grasping policy,'' in \emph{2021 IEEE International Conference on Robotics and Automation (ICRA)}.\hskip 1em plus 0.5em minus 0.4em\relax IEEE, 2021.

\bibitem{wu2024cross}
R.~Wu, T.~Zhu, X.~Lin, and Y.~Sun, ``Cross-category functional grasp transfer,'' \emph{IEEE Robotics and Automation Letters}, 2024.

\bibitem{madry2012object}
M.~Madry, D.~Song, and D.~Kragic, ``From object categories to grasp transfer using probabilistic reasoning,'' in \emph{2012 IEEE International Conference on Robotics and Automation}, 2012.

\bibitem{khargonkar2022neuralgrasps}
N.~Khargonkar, N.~Song, Z.~Xu, B.~Prabhakaran, and Y.~Xiang, ``Neuralgrasps: Learning implicit representations for grasps of multiple robotic hands,'' in \emph{Conference on Robot Learning (CoRL)}, 2022.

\bibitem{nvidia2023-isaac-sim}
\BIBentryALTinterwordspacing
NVIDIA, ``Nvidia isaac sim: Robotics simulation and synthetic data,'' 2023. [Online]. Available: \url{https://developer.nvidia.com/isaac-sim}
\BIBentrySTDinterwordspacing

\bibitem{jiang2021hand}
H.~Jiang, S.~Liu, J.~Wang, and X.~Wang, ``Hand-object contact consistency reasoning for human grasps generation,'' in \emph{Proceedings of the IEEE/CVF international conference on computer vision}, 2021.

\bibitem{qi2017pointnet++}
C.~R. Qi, L.~Yi, H.~Su, and L.~J. Guibas, ``Pointnet++: Deep hierarchical feature learning on point sets in a metric space,'' \emph{Advances in neural information processing systems}, vol.~30, 2017.

\bibitem{liu2022gdrnpp_bop}
X.~Liu, R.~Zhang, C.~Zhang, B.~Fu, J.~Tang, X.~Liang, J.~Tang, X.~Cheng, Y.~Zhang, G.~Wang, and X.~Ji, ``Gdrnpp,'' \url{https://github.com/shanice-l/gdrnpp_bop2022}, 2022.

\bibitem{deng2019pose}
X.~Deng, A.~Mousavian, Y.~Xiang, F.~Xia, T.~Bretl, and D.~Fox, ``Poserbpf: A rao-blackwellized particle filter for 6d object pose tracking,'' in \emph{Robotics: Science and Systems (RSS)}, 2019.

\bibitem{romero2017embodied}
J.~Romero, D.~Tzionas, and M.~J. Black, ``Embodied hands: modeling and capturing hands and bodies together,'' \emph{ACM Transactions on Graphics (TOG)}, vol.~36, no.~6, pp. 1--17, 2017.

\bibitem{pavlakos2024reconstructing}
G.~Pavlakos, D.~Shan, I.~Radosavovic, A.~Kanazawa, D.~Fouhey, and J.~Malik, ``Reconstructing hands in 3d with transformers,'' in \emph{Proceedings of the IEEE/CVF Conference on Computer Vision and Pattern Recognition}, 2024, pp. 9826--9836.

\end{thebibliography}
